\def\year{2019}\relax
\documentclass[letterpaper]{article} 
\usepackage{aaai19}  
\usepackage{times}  
\usepackage{helvet}  
\usepackage{courier}  
\usepackage{url}  
\usepackage{graphicx}  
\usepackage{filecontents}

\makeatletter
    \let\@internalcite\cite
    \def\cite{\def\citeauthoryear##1##2{##1, ##2}\@internalcite}
    \def\shortcite{\def\citeauthoryear##1{##2}\@internalcite}
    \def\@biblabel#1{\def\citeauthoryear##1##2{##1, ##2}[#1]\hfill}
\makeatother
\usepackage{natbib}

\usepackage{times}
\usepackage{epsfig}
\usepackage{graphicx}
\usepackage{amsmath}
\usepackage{amssymb}

\usepackage{multirow}
\usepackage{tabularx}
\usepackage{dsfont}
\usepackage{url}
\usepackage{subfigure}
\usepackage{amsthm}
\usepackage[export]{adjustbox}
\usepackage{wrapfig}

\newtheorem{theorem}{Theorem}
\newtheorem{lemma}[theorem]{Lemma}

\theoremstyle{definition}

\theoremstyle{remark}

\begin{document}

%
\title{Non-Parametric Transformation Networks}
\author{Dipan K. Pal ~~~~~~~~~~~~~~~ Marios Savvides\\
Department of Electrical and Computer Engineering\\
Carnegie Mellon University\\
Pittsburgh, PA 15213
}

\maketitle
\begin{abstract}
ConvNets, through their architecture, only enforce invariance to translation. In this paper, we introduce a new class of deep convolutional architectures called Non-Parametric Transformation Networks (NPTNs) which can learn \textit{general} invariances and symmetries directly from data. NPTNs are a  natural generalization of ConvNets and can be optimized directly using gradient descent. Unlike almost all previous works in deep architectures, they make no assumption regarding the structure of the invariances present in the data and in that aspect are flexible and powerful. We also model ConvNets and NPTNs under a unified framework called Transformation Networks (TN), which yields a better understanding of the connection between the two. We demonstrate the efficacy of NPTNs on data such as MNIST with extreme transformations and CIFAR10 where they outperform baselines, and further outperform several recent algorithms on ETH-80. They do so while having the same number of parameters. We also show that they are  more effective than ConvNets in modelling symmetries and invariances from data, without the explicit knowledge of the added arbitrary nuisance transformations. Finally, we replace ConvNets with NPTNs within Capsule Networks and show that this enables Capsule Nets to perform even better. 
\end{abstract}

\noindent

\section{Introduction}



\begin{figure}
\centering
\begin{tabular}{c}

        \subfigure[A single NPTN node]{%
        \centering
           \includegraphics[width=0.65\columnwidth,valign=m]{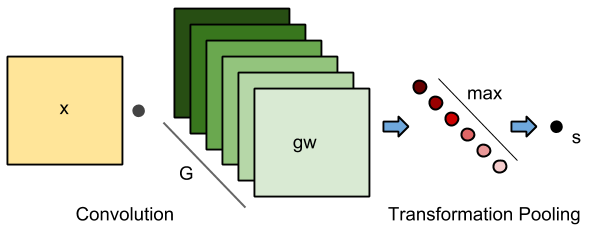}
\label{fig_nptn_node}
        }\\
         \subfigure[Types of Invariances in Deep Networks]{%
        \centering
    \includegraphics[width=0.65\columnwidth,valign=m]{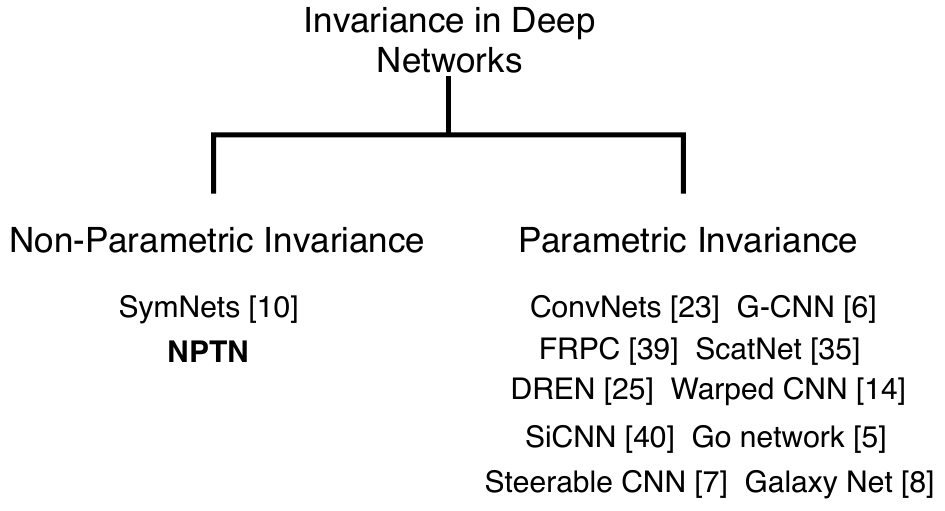}\label{fig_tree}
        }

 
\end{tabular}

    \vspace{-0.0cm}
\caption{ (a) Operation performed by a single Transformation Network (TN) node (single channel input and single channel output, Non-Parametric Transformation Networks are a kind of TN). TNs (and NPTNs) are a generalization of ConvNets towards learning general invariances and symmetries. The node has two main components (i) Convolution and (ii) Transformation Pooling. The dot-product between the input patch (x) and a set of $|G|$ number of filters $gw$ (green) is computed (this results in convolution when implemented with spatially replicated nodes). Here $|G|=6$ (different shades of green indicate transformed templates). $g$ indicates the transformation applied to the template or filter $w$ . The resultant six output scalars (red) are then max-pooled over to produce the final output $s$ (black). The pooling operation here is not spatially (as in vanilla ConvNets) but rather across the $|G|$ channels which encode non-parametric transformations. The output $s$ is now invariant to the transformation encoded by the set of filters $G$. Each plane indicates a single feature map/filter.  (b) Most previous works in deep learning have focused on invariances to transformations that are parametric in nature and fixed. SymNets \cite{gens2014deep} and NPTNs to the best of our knowledge are the only architectures to \textit{learn} invariances from data towards transformations that are not modelled by any expression in the network itself, \emph{i.e.} the symmetries that are captured are non-parametric in nature.} 
\vspace{-0.7cm}
\end{figure}

\textbf{The Fundamental Problem. }One of the central problems of machine learning, has been supervised classification. A core challenge towards these problems is the encoding or learning of invariances and symmetries that exist in the training data.  Indeed, methods which incorporate some known invariances or promote learning of more powerful invariances for a learning problem perform better in the target task given a certain amount of data. A number of ways exist to achieve this. One can present transformed versions of the training data \cite{burges1997improving,niyogi1998incorporating}, minimize auxiliary objectives promoting invariance during training \cite{schroff2015facenet,hadsell2006dimensionality} or pool over transformed versions of the representation itself \cite{liao2013learning,pal2016discriminative,pal2017mmif}.

\textbf{Convolutional Networks and Beyond. }  Towards this goal, ideas proposed in \cite{lecun1998gradient} with the introduction of convolutional neural  networks have proved to be very useful. Weight sharing implemented as convolutions followed by pooling  resulted in the hard encoding of translation invariances (and symmetries) in the network. This made it one of the first applications of modelling invariance through a network's architecture itself. Such a mechanism resulted in  greater regularization in the form of a structural or inductive bias in the network. With this motivation in mind, it is almost natural to ask whether networks which model more complicated invariances and symmetries perform better? Investigating architectures which invoke invariances not implicitly through the model's functional map but \textit{explicitly} through an architectural property seems important.

\textbf{New Dimensions in Network Architecture. } Over the years, deep convolutional networks (ConvNets) have enjoyed a wide array of improvements in architecture. It was observed early on that a larger number of filters (width) in ConvNets led to improved performance, though with diminishing returns. Another significant milestone was the development and maturity of residual connections and dense skip connections \cite{he2016deep,huang2016densely}. Though there have been more advances in network architecture, many of the improvements have been derivatives of these two ideas \cite{zagoruyko2016wide,chen2017dual,hu2017squeeze}. Recently however, Capsule Nets were introduced \cite{sabour2017dynamic} which presented another potentially fundamental idea of encoding properties of an entity or an object in an activity vector rather than a scalar. With the goal of designing more powerful networks, ideas presented in this paper for modelling \textit{general invariances in the same framework as ConvNets}, open up a new and potentially key dimension for architecture development.






\textbf{Primary Contribution.} In this work, we explore one such architecture class, called Transformation Networks (TN) which is a generalization of ConvNets. Additionally, we introduce a new type of TN using which a new class of networks can be built called Non-Parametric Transformation Networks (NPTNs). NPTNs networks have the ability to \textit{learn}  invariances to general transformations that are \textit{observed} in the data which are non-parametric in nature (difficult to express mathematically). They can be easily implemented using standard off-the-shelf deep learning frameworks and libraries. Further, they can be optimized using vanilla gradient descent methods such as SGD. Unlike other methods that enforce additional invariances in convolutional architectures \cite{teney2016learning,wu2015flip,li2017deep}, NPTNs do not need to transform the input, activation maps or the filters at any stage of the learning/testing process. They enjoy benefits of a standard convolutional architecture such as speed and memory efficiency while being more powerful in modelling invariances and being elegant in their operation.   When forced to ignore any learnable transformation invariances in data, they gracefully reduce to vanilla ConvNets in theory and practice. However, when allowed to do so, they outperform ConvNets by capturing more general invariances.


\textbf{Some properties of NPTNs.} The architecture itself of an NPTN allows it to be able to learn powerful invariances from data provided the transformations are observable in data (a single node is illustrated in Fig.~\ref{fig_nptn_node}). NPTNs do not enforce any invariance that is not observed in the data (although translation invariance can still be enforced through spatial pooling). Learning invariances from data is different and more powerful than enforcing known and specific invariances as is more common in literature (see Fig.~\ref{fig_tree}). Networks which enforce predefined symmetries (including vanilla ConvNets) force the same invariances at all layers which is a strong prior. More complex invariances are left for the network to learn using the implicit functional map as opposed to the explicit architecture. The proposed NPTNs have the ability to learn \textit{different} and independent invariances for different layers and in fact for different channel paths themselves. Vanilla ConvNets enforce translation invariance through the convolution operation followed by a aggregation operation (either pooling or a second convolution layer) and only need to learn the \textit{filter instantiation}. However, an NPTN node needs to learn 1) the instantiation of the filter and 2) the transformation that the particular node is invariant towards encoded as a \textit{set} of filters. Each node learns these entities independently of each other which allows for a  more flexible invariance model as opposed to architectures which replicate invariances across the network.


\section{Prior Art}


Although past applications of incorporating invariances  were more specific and relatively narrow, development of such methods offers a better understanding of the importance of the problem.  Though in this work we focus on deep architectures, it is important to note a number of works on modifications of Markov Random Fields and Restricted Boltzman Machines to achieve rotational invariance \cite{schmidt2012learning,kivinen2011transformation,sohn2012learning}.

\textbf{Incorporating known invariances using deep networks. } Convolutional architectures have seen many efforts to produce rotation invariant representations. \cite{fasel2006rotation} and \cite{dieleman2015rotation} rotate the input itself before feeding it into stacks of CNNs and generating rotation invariant representations through gradual pooling or parameter sharing.  \cite{teney2016learning,wu2015flip,li2017deep} rotate the convolution filters (a cheaper albeit still expensive operation) instead of transforming the input followed by pooling. A similar approach was explored for scale by \cite{xu2014scale} and for Go by \cite{clark2015training}. An interesting direction of research was explored by \cite{sifre2013rotation} where the rotation, scale and translation invariant filters were fixed and non-trainable.  \cite{cohen2016group,henriques2016warped} presented methods to incorporate parametric invariances using groups and warped convolutions.  The transformations in  \cite{henriques2016warped,cohen2016steerable} are known apriori and the sample grids and steerable filters are generated offline. This limits the capability to learn arbitrary and adaptive transformations. NPTNs need no such apriori knowledge apart that enocded in its architecture, can learn arbitrary non-parametric transformations and finally are simpler and more elegant  in implementation.

\textbf{Learning unknown invariances from data.} 
In most real world problems, nuisance transformations present in data are unknown or too complicated to be parameterized by some function. \cite{anselmi2013unsupervised} proposed a theory of group invariances called I-theory and explored its connection to general classification problems and deep networks. Based off the core idea of measuring moments of a group invariant distribution, multiple works had demonstrated efficacy of the ideas in more challenging real-world problems such as face recognition, though not in a neural network setting \cite{liao2013learning,pal2016discriminative,pal2017mmif}. 


\textbf{Learning unknown invariances from data using deep networks.} Very few works have explored incorporating unknown invariances into deep networks. To the best of our knowledge, SymNets \cite{gens2014deep} was only one other previous study proposed deep networks which \textit{learn} more general transformations. They were introduced as one of the first to model general invariances with back propagation. They utilize kernel based interpolation to tie weights enable them to model general symmetries. Nonetheless, the approach is complicated and difficult to scale. \cite{anselmi2017symmetry} provide sufficient conditions to enforce the learned representation to have symmetries learned from data. \cite{kavukcuoglu2009learning} modelled local invariances using pooling over sparse coefficients of a dictionary of basis functions. \cite{ngiam2010tiled} achieved local invariance through complex weight sharing. Optimization was carried out through Topographic ICA and only carried out layer wise for deep networks. A separate approach towards modelling invariances was also developed where a normalizing transformation is applied to every input independently. This approach was applied to transforming auto encoders \cite{hinton2011transforming} and Spatial Transformer Networks \cite{jaderberg2015spatial}.

\section{The Transformation Network Paradigm}

A Transformation Network (TN) is a feed forward network with its architecture designed to enforce invariance to some class of transformations through pooling. At the core of the framework is the TN node.  A TN network consists of multiple such nodes stacked in layers. A single TN node is analogous to a single convolution layer with single channel input and single channel output. 


Each TN node (single input channel and single output channel) internally consists of two operations 1) \textit{(convolution)} the convolution operation with a bank of filters  and 2) \textit{(transformation pooling)} a max pooling operation \textit{across} the set of the resultant convolution feature maps from the single input channel.  Note the pooling is not spatial but rather across channels originating from the \textit{same input channel} (this is different from MaxOut \cite{goodfellow2013maxout} which pools over all input channels\footnote{We discuss deviation from MaxOut in more detail in the supplementary.}). Fig.~\ref{fig_nptn_node} illustrates the operation of single TN node with a single input/output channel for a single patch. The single channel illustrated in the figure takes in a single input feature map and convolves it with a bank of $|G|$ filters. Here $|G|$ is the cardinality (or size) of the set of transformations that the TN node is invariant towards, with $G$ being the actual set itself. Next, the transformation max pooling operation max pools across the $|G|$ feature values to obtain a single TN activation value. When this node is replicated spatially, standard convolution layers can be utilized. Formally, a TN node denoted by $\Upsilon$  acting on a 2D image patch vectorized as $x\in \mathbb{R}^d$ can be defined as follows.
\begin{align}
  \Upsilon(x) &= \max_{g \in G} ( \langle  x, gw  \rangle )  \label{TN_eq_1}
\end{align}
Here, $\langle~~~\rangle$ denotes a dot product and $G$ is formally defined as a unitary group, \emph{i.e.} a finite set obeying group axioms with each element being unitary. $w\in \mathbb{R}^d$ is the weight or filter, and $gw$ is the group element $g$ acting on $w$\footnote{We use this shorter notation to reduce clutter.}. Therefore, the convolution kernel weights of a TN node are simply the transformed versions of $w$ as transformed by the unitary group $G$. The TN node has to, only in theory, transform weight template $w$ according to $G$ to generate the rest of the filters to be pooled over during the transformation pooling stage. In practice however, these are simply stored as a set of templates or filters which only \textit{implicitly} encode $G$ through some constraints. For instance, vanilla ConvNets model the group $G$ to be the translation group by enforcing it through the convolution operation. Thus, \textit{a ConvNet can be exactly modelled by the TN framework when $G$ is the translation group.} Gradient descent updates the filter $w$ for a single node which immediately specifies the other filters in that node since they are the translated versions of $w$.


\textbf{Significance of the Modeling of Transformations as Unitary Groups.} The use of unitary groups to model transformations and invariances has emerged as a prominent theoretical tool \cite{anselmi2013unsupervised,pal2017mmif}. Group structure allows the computing of  invariant objects such as group integrals. However, the significance of the unitary group lies in the fact that the vanilla ConvNet is invariant to translations, which is the simplest unitary group. Any framework that models invariance using the unitary group can be directly generalized to more complex groups such as rotations (rotation is an unitary transformation). This allows for seamless integration of the vanilla ConvNet into the theoretical framework and provides clear theoretical and practical connections to the same. Unitary groups in TNs allow them to exactly model ConvNets while generalizing to more complex networks invoking more complex invariances. The unitary group condition thus is only a useful theoretical tool, however should not be considered as a practical constraint.


\textbf{Invariances in a TN node.} Invariance in the TN node  arises directly due to the symmetry of the unitary group structure of the filters. The max operation simply measures the infinite moment of an invariant distribution which invokes invariance. We demonstrate this in the form of the following simple result\footnote{Proof in the supplementary.}.

\begin{lemma}\label{lem_invariance} (Invariance Property) Given vectors $x, w \in \mathbb{R}^d$, a unitary group $\mathcal{G}$ and  $\Upsilon(x) = \max_{g \in G} ( \langle  x, gw  \rangle )$, for any fixed $ g' \in \mathcal{G}$, then  $\Upsilon(x) =  \Upsilon(g'x) $.

\end{lemma}

Lemma~\ref{lem_invariance} shows that for \textit{any} input $x$ (including test inputs), the node output is \textit{invariant} to the transformation group $G$. Note that invariance to test samples arises from two components. First, the group structure of $G$ provides exact invariance and second, the unitary condition allows for the invariance properties to be extended to unseen test samples. This is interesting, since one does not need to observe any transformed version of say a test sample $x$ during training which reduces sample complexity \cite{anselmi2013unsupervised}. Invariance is invoked for any arbitrary input $x$ during test time, thereby demonstrating good generalization properties.

\textbf{Relaxing towards Non-group and Non-Unitary Structure in a TN node (Towards NPTNs). } Lemma~\ref{lem_invariance} guarantees exact invariance perfectly for vanilla ConvNets and TNs which model $G$ as having a group-structure and the unitary condition. For methods that do not enforce these conditions (unitary group conditions) in theory, no test generalization claim can be made. However, a number of studies have observed approximate albeit sufficient invariances in practice under this setting \cite{anselmi2013unsupervised,anselmi2017symmetry,pal2016discriminative,pal2017mmif,liao2013learning}. The main motive for modelling transformations as unitary groups was to provide a theoretical connection to ConvNets and other methods that enforce other kinds of unitary invariance such as rotation invariance \cite{li2017deep,wu2015flip}. However, real-world data experiences a large array of transformations acting, which certainly lie outside the span of unitary transformations. Keeping this in mind, constraining the network to model only unitary transformations limits their ability to learn these more general invariances which are difficult to characterize. 

In the following section, we introduce a new kind of TN called the NPTN which is free from the constraints and limitations of unitary modelling, thereby being more expressive. Indeed, in our experiments, we observe that the NPTN architectures are able to perform better by learning invariance (signified by better test generalization) towards both 1) group structured, unitary and parametric transformations such as translations and rotations, and also towards 2) general non-group structured and non-parametric transformations (as in general object classification) which are difficult to characterize. Note that Lemma~\ref{lem_invariance} only serve as a result for ConvNets and TNs, they do not characterize the invariance properties on NPTNs and general non-group non-unitary transformation. Investigation of such properties of NPTNs under the general setting is arduous and is outside the scope of this paper. Further, note that developing TNs and relating the unitary condition is not necessary for the development or motivation of NPTNs. TNs however provide a more elegant story and more importantly clarify the connection to vanilla ConvNets and helps to put our contribution in perspective.

\section{Non-Parametric Transformation Networks}\label{sec_NPTN}


A Non-Parametric Transformation Network (NPTN) is a kind of TN that lacks any constraints on set of weights/filters $w$ for any particular node. Here the set of filters $G$ has two relaxations 1) need not have any group structure and 2)  need not model any parametric and/or unitary transformations such as the translations or rotations. The term $G$ in an NPTN represents simply a \textit{set} of arbitrary filters modelling arbitrary transformations which are (potentially) non-parametric. One might think of the analogy from statistics where the Gaussian distribution is parametric, however for many real-world distributions a non-parametric tool such as a histogram is more appropriate. Note that however, there is no constraint that prevents a NPTN from learning translation and rotation invariance. In fact, in one of our experiments this is exactly the requirement.  \textit{Under the two relaxations, the invariance invoked to these arbitrary transformations in an NPTN would only be approximate.} Nonetheless and consistent with previous work, we find in our experiments that despite the approximation, there is much to be gained overall and the invariance invoked suffices in practice as also found by  \cite{liao2013learning,pal2016discriminative}. 

\begin{figure*}
    \begin{center}
        \subfigure[Convolution layer]{%
        \centering
            \includegraphics[width=0.55\columnwidth,valign=m]{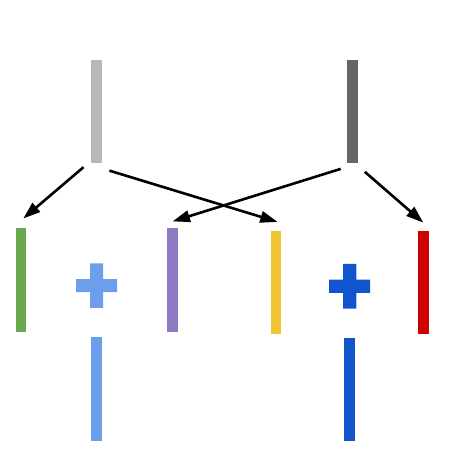}\label{fig_1_a}
        }%
         \subfigure[NPTN layer]{%
        \centering
        \includegraphics[width=0.55\columnwidth,valign=m]{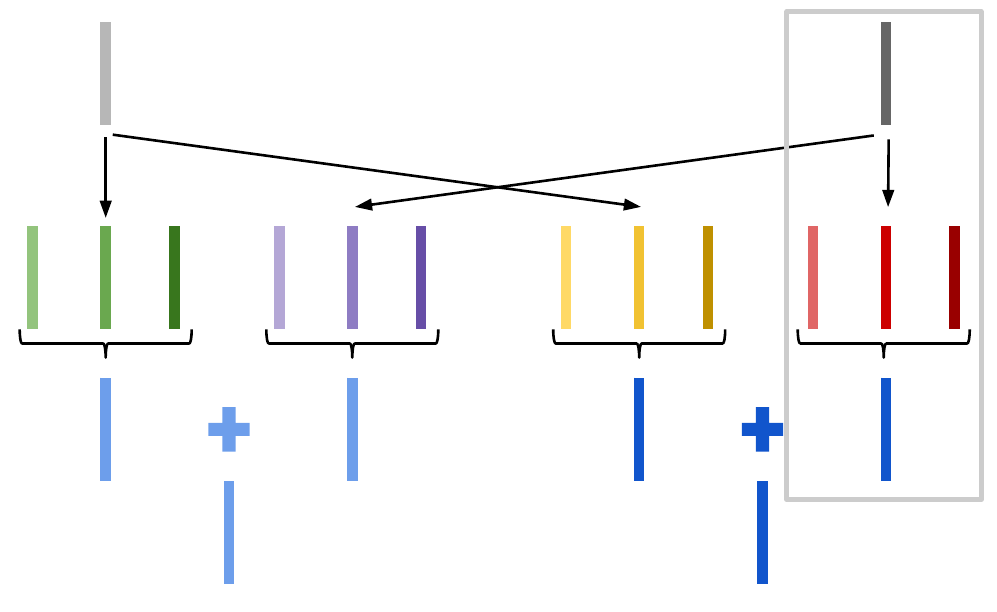}\label{fig_1_b}
        }
     \subfigure[Relation between ConvNets and NPTNs]{%
        \centering
        \includegraphics[width=0.75\columnwidth,valign=m]{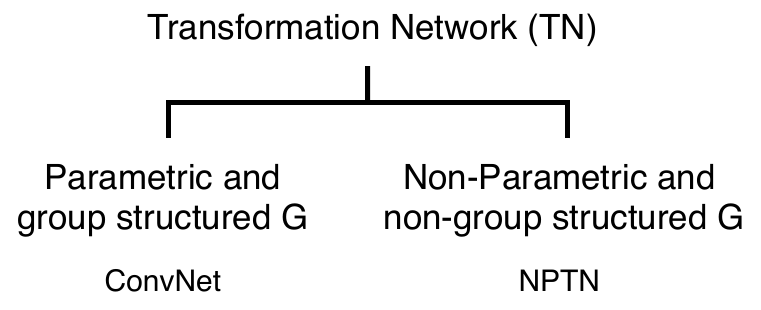}\label{fig_1_c}
        }
    \end{center}
    \vspace{-0.5cm}
\caption{Comparison between (a) a standard Convolution layer and (b) a NPTN layer with $|G|=3$. Each layer depicted has 2 input (shades of grey) and 2 output channels (shades of blue). The light grey rectangle encloses a single TN node (see Fig.~\ref{fig_nptn_node}) The convolution layer has therefore, $2\times 2=4$ filters, whereas the NPTN layer has $2\times 2\times 3=12$ filters. The different shades of filters in the NPTN layer denote transformed versions of the same filter (same color) which are max pooled over (support denoted by inverted curly bracket). The + operation denotes channel addition. In our experiments, we adjust the input/output channels of the NPTN layer to have the same number of parameters as the ConvNet baselines. (c) Shows how ConvNets and NPTNs are categorized under the TN framework. }
\label{fig_conv_nptn}
    \vspace{-0.5cm}

\end{figure*}

In an NPTN, both the entities ($w, G$) are learned, \emph{i.e.} a NPTN node is tasked with learning both the filter instantiation $w$, \textit{and} the set of transformations $G$ to which the node is to be invariant towards. Nonetheless and rather importantly, no generation of transformed filters is necessary during any forward pass of a NPTN layer since the set $G$ of transformed filters is always maintained and updated by gradient descent. This significantly reduces computational complexity compared to some previous works \cite{teney2016learning,wu2015flip}. Learning $G$ from data is in sharp contrast with the vanilla convolutional node in which only the filter instantiation $w$ is learned and where $G$ is \textit{hard coded} to be the translation group which is a parametric transformation (and also arguably the most elementary). Thus, ConvNets are a kind of Parametric Transformation Networks (PTNs) (see Fig.~\ref{fig_1_c}). It is also important to note that however, setting $|G|=1$ \textit{and} incorporating spatial pooling, a NPTN is reduced to a vanilla ConvNet in practice. Compared to other approaches to learn and model general invariances such as SymNets \cite{gens2014deep}, the NPTN architecture is elegantly simple and also a close generalization of ConvNets.  Further, they can replace any convolution layer in any architecture making them versatile. We now describe the NPTN layer in more detail and discuss its characteristics.



\textbf{NPTN Layer Structure, Forward Pass and Training.} Fig.~\ref{fig_conv_nptn} illustrates a NPTN layer and compares it to a vanilla ConvNet layer. The NPTN layer shown has 2 input channels, 2 output channels and $|G|=3$.  For a NPTN layer with $M$ input channels and $N$ output channels, there would be $MN$ NPTN nodes each identical to the one shown in Fig.~\ref{fig_nptn_node}. There are $|G|$ filters learned for each of the $MN$ nodes, which each are convolved over the image similar to a vanilla ConvNet.  Consider Fig.~\ref{fig_1_b}, once the input is convolved with the $M \times |G|$ filters, the $M$ sets each with $|G|$ feature maps each are max pooled across the $|G|$ feature maps. More specifically, each of $|G|$ feature maps from a single input channel results in one intermediate feature map after max pooling (across the $|G|$ channels). This is the primary step that invokes invariances to transformations. After this operation there are $MN$ intermediate feature maps which are transformation invariant. Now, the sum (alternatively the mean) of these $M$ feature maps results in one \textit{output} feature map or channel. This is repeated for each of the $N$ output channels\footnote{We provide implementation details of NPTNs using standard libraries in the supplementary.
}. Note that there is no operation in this forward pass where the input or the filters need to be transformed on-the-fly, which makes it NPTNs computationally efficient compared to some previous models  \cite{fasel2006rotation,dieleman2015rotation,teney2016learning,wu2015flip,li2017deep}. In fact, the computation complexity for NPTNs only increases with the order $|G|$ relative to a vanilla convolution layer. This is countered in our experiments by decreasing $M$ and $N$, primarily to preserve the number of parameters. The NPTN layer can be trained using standard back-propagation. Back-propagation updates each of the $|G|$ filters of the NPTN independently depending on which of the $|G|$ filters is the `winner' during the channel max pooling operation. Note again that this operation is very different from MaxOut which pools over inputs from \textit{all} channels, whereas here each max operation pools over $|G|$ channels only from the \textit{same input} channel\footnote{We discuss deviation from MaxOut in more detail in the supplementary.}. Since the filters are not constrained to form any group, we do not expect to see any regular transformations being observed in the filters (for instance, rotated filters for rotation invariance). This might seem as a slight hindrance to interpretibility, nonetheless in our experiments, we find NPTNs perform well in specific applications where learning invariance from the data is necessary.


\textbf{Invariance Modelling in NPTNs is Data Driven and Highly Flexible.}  It is important to note that though the architecture of NPTNs allows it to \textit{learn} invariances, it does \textit{not} in fact enforce any particular invariance by itself. NPTNs can only learn invariances to transformations that are observed in data, and thereby are even more benefited from data augmentation and natural variation. This is a critical difference between NPTNs and other works which do enforce specific invariances through design (see under Parametric Invariance in Fig.~\ref{fig_tree}). Another important and powerful property that emerges from having independent filter sets for each of the NPTN nodes in an entire network, is that each individual node can model invariance to a completely different transformation. Concretely, a single NPTN layer with $M$ input channels and $N$ output channels potentially can model $MN$ different kinds of invariances. This is again in sharp contrast to ConvNets and other previous works such as \cite{teney2016learning,wu2015flip,li2017deep} where each layer and in fact each of the channel paths model exact same invariance, either translation, rotation or scale. NPTNs thus offers immense flexibility in invariance modelling.

\section{Empirical Evaluation of NPTNs}

\begin{figure}
\centering
\includegraphics[width=0.8\columnwidth,valign=m]{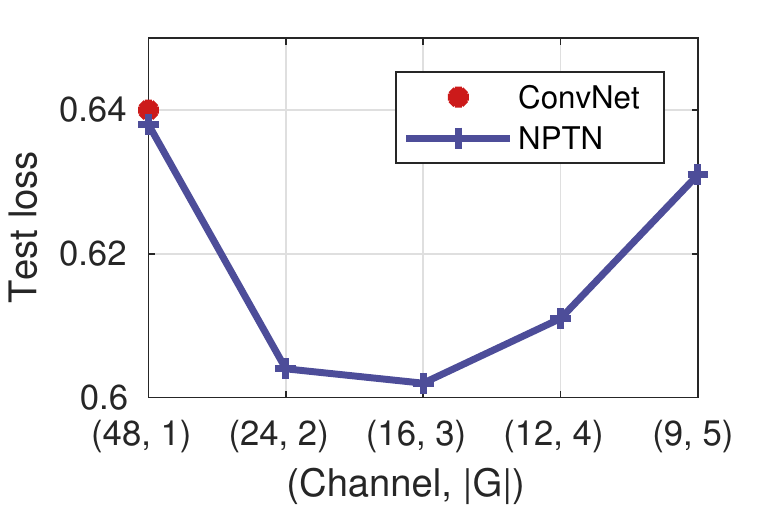}
    \vspace{-0.0cm}
\caption{Test losses on CIFAR10 for the two layered network. Each network listed has the same number of filters.} 
\label{fig_bench_2L}
\end{figure}

\subsection{Benchmarking against ConvNets on CIFAR10}

In our first set of experiments, we benchmark and characterize the behavior of NPTNs against the standard ConvNets augmented with Batch Normalization \cite{ioffe2015batch}. The goal of this set of experiments is to observe whether learning non-parametric transformation invariance from complex visual data itself helps with object classification. For this experiment, we utilize the CIFAR10 dataset\footnote{With standard data augmentation of random cropping after a 4 pixel pad, and random horizontal flipping. Training was for 300 epochs with the learning rate being 0.1 and decreased at epoch 150, and 225 by a factor of 10.}. The networks we experiment with are not designed to compete with state-of-the-arts on this data but rather throw light into the behavior of NPTNs. We therefore utilize a small network, specifically a two layered network, for these experiments. Each layer block of the baseline ConvNets consist of the convolution layer, followed by batch normalization and the non-linearity (PReLU) and finally by a 2 by 2 spatial max pooling layer. Each corresponding NPTN network replaces only the convolution layer with the NPTN layer. Thus, NPTN is allowed to model non-parametric invariance in addition to the typically enforced translation invariance due to spatial max pooling. The two layered network baseline ConvNet has channels [3, 48, 16] with a total of $3 \times 48 + 48 \times 16 = 912$ filters. The NPTN variants in this experiment keep the total number of filters  constant with 48 channels with $|G|=1$ denoted by (48 1), 24 channels with $|G|=2$ denoted by (24 2), and so on up until 9 channels with $|G|=5$ (9 5).  Fig.~\ref{fig_bench_2L} shows the testing losses. Each network experimented with has the same number of parameters. We find all NPTN variants which learn a non-trivial set of transformations ($|G|>1$) outperform the ConvNet baseline significantly, with NPTN $|G|=3$ performing the best. 
\begin{table}
\centering
\begin{tabular}{l c } 
\hline
\hline 
Method    &   Accuracy (\%)    \\
\hline
\hline
ConvNet \cite{khasanova2017graph}  &    80.1    \\
STN  \cite{jaderberg2015spatial}    &    45.1   \\
DeepScat \cite{oyallonroto2015}    &   87.3    \\
   HarmNet \cite{worrall2017harmonic}   &    94.0  \\
  TIGradNet \cite{khasanova2017graph}    &   95.1    \\
\hline
NPTN (Ours)      &   \textbf{96.2}  \\
\hline
\end{tabular}
\caption{Test accuracy on ETH-80. All models including NPTNs and the ConvNet had roughly the same number of parameters (about 1.4M). Models followed the architecture described in \cite{khasanova2017graph}. Results for models other than NPTN are cited as is from \cite{khasanova2017graph}. }
\label{tab_exp_eth}
\end{table}


\subsection{Benchmarking against other approaches: ETH-80} We now benchmark NPTNs against other approaches learning invariances on the ETH-80 dataset \cite{leibe2003analyzing}. As our baseline, we follow the experimental setup and the specifications of the models described in \cite{khasanova2017graph}. Note that for this experiment, our goal is not to attain state-of-the-art results, but rather benchmark against other related methods under a comparable setting. The dataset has 80 objects belonging to 8 classes. Each object has 41 images taken from a grid of different viewpoints on a hemisphere. Following \cite{khasanova2017graph}, we resize the images to $50\times 50$ and train on 2,300 images and test on the rest. The isometric transformations in the dataset present a good challenge for approaches to invoke invariance in a real-world setting. For this experiment, we compare against standard ConvNets, Spatial Transformer Networks \cite{jaderberg2015spatial}, DeepScat \cite{oyallonroto2015}, HarmNet \cite{worrall2017harmonic} and TIGradNet \cite{khasanova2017graph}. The NPTN architecture was chosen to by replacing the convolution layers in the ConvNet architecture in \cite{khasanova2017graph} with NPTN layers while setting $|G|=3$ and reducing the number of channels to preserve the number of parameters. All models in this experiment (including NPTNs) have about 1.4M parameters. Table~\ref{tab_exp_eth} presents the test accuracy on ETH-80. We find that NPTN outperforms these other high-performing algorithms on this task with an accuracy of 96.2 \%. Thus, NPTNs despite having much simpler architecture and the same number of parameters, is able to perform well in a task where the primary nuisance transformation is due to varying 3D pose of the objects.

\subsection{Learning Unknown Transformation Invariances from Data}\label{exp_transrot_inv}

\begin{table}
\centering
\begin{tabular}{l c c c c} 

\hline\hline 
Rotations   &     $0^\circ$ &   $30^\circ$  &     $60^\circ$  &$90^\circ$  \\
\hline
\hline
ConvNet (36) &   0.75   &   1.16    &  2.05  &    3.32    \\
\hline

NPTN (36, 1) &    0.68     &   1.27     &   2.01   &     3.36    \\
NPTN (18, 2) &    0.66  &  1.09 )     &  1.72     &     2.88 \\
NPTN (12, 3) & \textbf{ 0.63}      &  \textbf{  1.08 }  &   \textbf{1.71 }  &    \textbf{ 2.76 } \\
NPTN (9, 4) &   0.66    &  1.17   & 1.83  &  2.94   \\
\hline
\hline
Translations   &     0 pix &  4 pix  &    8 pix   & 12 pix  \\

\hline
ConvNet (36) &  0.62   & 0.95     &  1.97   &  7.00 \\
\hline
NPTN (36, 1) &  \textbf{ 0.62}   &  0.88   &  1.84    &  7.22       \\
NPTN (18, 2) &  0.74 &  0.75  &  1.70   &  6.26 \\
NPTN (12, 3) &    0.66  &  \textbf{ 0.70}  &  \textbf{ 1.58}   &  \textbf{ 6.20} \\
NPTN (9, 4) &   0.64  &  0.76  &   1.59    &  6.37  \\


\hline
\end{tabular}
\caption{Test error on progressively transformed MNIST with (a) random rotations and (b) random pixel shifts. NPTNs can learn invariances to arbitrary transformations from the data itself without any a priori knowledge. All models have same number of parameters. }
\label{tab_exp_trans_1}
    \vspace{-0.5cm}

\end{table}

We now demonstrate the ability of NPTN networks to learn invariances directly from data without any apriori knowledge. For this experiment, we augment MNIST with extreme a) random rotations b) random translations, \textit{both} in training and testing data thereby increasing the complexity of the learning problem itself.  For each sample, a random instantiation of the transformation was applied. For rotation, the angular range was increased, whereas for translations it was the pixel shift range. Table~\ref{tab_exp_trans_1} presents these results. All networks in the table are two layered and have the exact same number of parameters. As expected, NPTNs match the performance of vanilla ConvNets when there were no additional transformations added ($0^\circ$ and $0$ pixels)\footnote{NPTNs perform slightly better than ConvNets for $0^\circ$ rotations because for all rotation experiments, small translations up to 2 pixels were applied only in training.}. However, as the transformation intensity (range) is increased, NPTNs perform significantly better than ConvNets. Trends consistent with previous experiments were observed with the highest performance observed with NPTN ($|G|=3$). This highlights the main feature of NPTNs, \emph{i.e.} their ability to model arbitrary transformations observed in data without any apriori information and without changes in architecture whatsoever. They exhibit better performance in settings where both rotation invariance and \textit{stronger} translation invariance is required (even though ConvNets are designed specifically to handle translations). This ability is something that previous deep architectures did not possess nor demonstrate.

\begin{figure}
\centering
\includegraphics[width=0.8\columnwidth,valign=m]{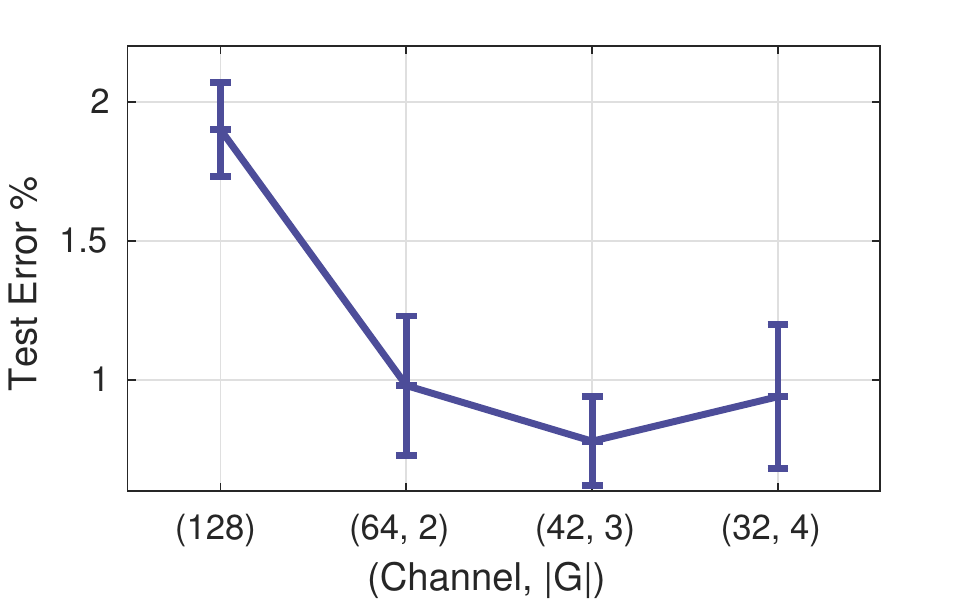}
    \vspace{-0.0cm}
\caption{Test errors on MNIST for Capsule Nets augmented with NPTNs. (128) denotes a Capsule Network with a vanilla ConvNet. Other labels are NPTNs with (channels, $|G|$). The number of filters from left to right is $\{ 4224, 4160,  4074,  4128 \}$. NPTNs significantly outperform ConvNets in Capsule Nets with fewer filters.} 
\label{fig_exp_capsnet}
\vspace{-0.5cm}
\end{figure}


\subsection{NPTNs with Capsule Networks}

Capsule Networks with dynamic routing were recently introduced as an extension of standard neural networks \cite{sabour2017dynamic}. Since the original architecture is implemented using vanilla convolution layers, invariance properties of the networks are limited.  Our goal for this experiment is to replace Convolution Capsule Nets with NPTN Capsules. We replace the convolution layers in the Primary Capsule layer of the published architecture with NPTN layers while maintaining the same number of parameters (by reducing number of channels and increasing $|G|$). Our baseline is the proposed CapsuleNet with 3 layers using a third party implementation in PyTorch\footnote{https://github.com/dragen1860/CapsNet-Pytorch.git}. The baseline convolution capsule layer had 128 output channels. The NPTN variants progressively decreased the number of channels as $|G|$ was increased. All other hyperparameters were preserved. The networks were trained on the 2-pixel shifted MNIST for 50 epochs with a learning rate of $10^{-3}$.  The performance statistics of 5 runs are reported in Fig.~\ref{fig_exp_capsnet}. We find that for roughly the same number of kernel filters (and parameters), Capsule Nets have much to gain from the use of NPTN layers (a significant test error decrease from 1.90 to 0.78 for $\frac{1}{3}$ of the baseline number of channels and $|G|=3$). The learning of invariances within each capsule significantly increases efficacy and performance of the overall architecture. 

\subsection{Current Limitations}

We observed that NPTNs and the idea of learning general non-parametric invariances form the data itself has potential and merits further investigation as a component in more broder studies. Nonetheless, our current implementation of NPTN in PyTorch suffers from high memory usage and high computation time partly due to channel shuffling. Given our limited computation resources as this time, the size and depth of the networks that we can train is subsequently limited. We therefore focus more on a systematic and thorough investigation and benchmarking of a smaller network on multiple tasks and experiments rather than larger scale experiments. Such studies will be within scope once more efficient function routines for specific tasks are available in deep learning packages. Nonetheless, even with the manageable size of our networks, NPTNs outperform several recent approaches on ETH-80 with the same number of parameters. Further, they can be incorporated into newer architectures such as Capsule Nets and improve performance.


\section{Discussion}

It is clear that the success of ConvNets is not the whole story towards solving perception. Studies into different aspects of network design will prove to be paramount in addressing the complex problem of not just visual  but general perception. 

The development of NPTNs offer one such design aspect, \emph{i.e.} learning non-parametric invariances and symmetries directly from data. Through our experiments, we found that NPTNs can indeed effectively learn general invariances without any apriori information. Further, they are effective and improve upon vanilla ConvNets even when applied to general vision data as presented in CIFAR10 and ETH-80 with complex unknown symmetries. This seems to be a critical requirement for any system that is aimed at taking a step towards general perception. Assuming detailed knowledge of symmetries in real-world data (not just visual) is impractical and successful models would need to adapt accordingly.

In all of our experiments, NPTNs were compared to vanilla ConvNet baselines with the same number of filters (and thereby more channels). Interestingly, the superior performance of NPTNs despite having fewer channels indicates that better modelling of invariances is a useful goal to pursue during design. Explicit and efficient modelling of invariances has the potential to improve many existing architectures. Indeed, we outperform several state-of-the-art algorithms on ETH-80. In our experiments, we also find that Capsule Networks which utilized NPTNs instead of vanilla ConvNets performed much better. This motivates and justifies more attention towards architectures and other solutions that efficiently model general invariances in deep networks. Such an endeavour might not only produce networks performing better in practice, it also promises to deepen our understanding of deep networks and perception in general.

\bibliographystyle{aaai}
\tiny{\bibliography{biblio}}


\end{document}